\documentclass[review]{elsarticle}

\usepackage{lineno,hyperref}
\usepackage{amsmath}
\usepackage{tabularx}
\usepackage{graphicx}
\usepackage{adjustbox}
\usepackage{geometry}

\journal{xxx}

\begin{document}

\begin{frontmatter}

\title{Fuzzy Win-Win: A Novel Approach to Quantify Win-Win Using Fuzzy Logic}


\author{Ahmad B. Hassanat}
\address{Faculty of information technology,\\ Mutah University \\ Karak, Jordan, hasanat@mutah.edu.jo}

\author{Ghada A. Altarawneh}
\address{Department of accounting,\\ Mutah University \\ Karak, Jordan}

\author{Ahmad S. Tarawneh}
\address{Department of algorithms and their applications,\\ Eotvos Lorand University \\Budapest, Hungary}

\author{David Carfì}
\address{Department of Mathematics,\\ University of California Riverside,  Riverside,  \\CA92521, USA}

\author{Abdullah Almuhaimeed}
\address{The National Centre for Genomics and Bioinformatics, \\King Abdulaziz City for Science and Technology, \\Riyadh, 11442, Saudi Arabia}

\begin{abstract}
The classic win-win has a key flaw in that it cannot offer the parties with right amounts of winning because each party believes they are winners. In reality, one party may win more than the other. This strategy is not limited to a single product or negotiation; it may be applied to a variety of situations in life. We present a novel way to measure the win-win situation in this paper. The proposed method employs the Fuzzy logic to create a mathematical model that aids negotiators in quantifying their winning percentages. The model is put to the test on real-life negotiation scenarios such as the Iraqi-Jordanian oil deal, and the iron ore negotiation (2005-2009). The presented model has shown to be a useful tool in practice and can be easily generalized to be utilized in other domains as well.

\end{abstract}


\end{frontmatter}

\nolinenumbers

\section{Introduction}
According to Merriam Webster dictionary, the first known use of the word win-win goes back to the year 1962. As per the dictionary, the concept is defined as “advantageous or satisfactory to all parties involved”.  In spite of the fact that some experts regard this concept as an oxymoron \cite{r1,r11,r12,r13}, it has captivated practically almost everyone on the planet and is used on a daily basis by almost everyone. Furthermore, it piqued academics interest in a wide range of topics and study disciplines, including Economy \cite{r1,r2,shaban2016sme, carfi2012coopetitive, dixon2021winners, xu2020can}, Business \cite{r3,r4,r5,r17,hassanat2015gambling}, Game Theory  \cite{r6,r7,r8,r9,r10,r35}, Sustainability \cite{r14,r15,r16},  Biology \cite{r18,r19,r20,r21,r22}, Policy \cite{r23,r24,r25}, Agriculture \cite{r26,r27}, Health \cite{r28,r29,r30}, Education \cite{r31,r32,r33,r34}, Social science \cite{r36,r37,r38,r39}, Engineering \cite{r40,r41,r42}, Tourism \cite{r43,r44}, etc.

The phrase "win-win" made its first documented appearance in Singer's work \cite{r48}, where he characterized it as follows: 
“In zero-sum games, every win for one side is a loss for the other; there can be no such thing as a ‘win-win’ outcome. The non-zero-sum game, on the other hand, is not one of pure conflict, but a mixture of conflict and co-operation. An excellent example is the “prisoner’s dilemma” noted in the previous chapter. Objectively, a win-win outcome was available, but the prisoners played as if it were not.”
Since then, the term has been defined several times based on the study discipline and the topic addressed; the following are examples of distinct definitions from various scholars:  

\begin{itemize}
    \item Covey defines win-win as “a frame of mind and heart that constantly seeks mutual benefit in business and personal transactions” \cite{r59}.
    \item Nalis et al. define win–win solutions as “outcomes of interpersonal behavior that exceed the outcomes that each participant could achieve alone” \cite{r51}.
    \item Fujita et al. define win-win as “ideas that might give both parties most of what they want” \cite{r55}.
    \item Thompson and Gonzales regard “win-win outcome to be one that is efficient- meaning that there is no other outcome parties could reach that at least one party would prefer without reducing the other party's utility” \cite{r56}.
    \item Brooks defines win–win as “success in two or more of the outcomes measured (ecological, economic, social) and “tradeoffs” are defined as some combination of success, limited success, or failure” \cite{r57}.
    \item Bottos, and Coleman define win-win in any negotiation as “the outcome that makes both parties feel as though they have benefited from the discussion” \cite{r58}.
    \item Carbonara et al. concluded that win-win is the capacity to meet the diverse interests of the parties involved by assuring their profit demands while also fairly allocating risk among them \cite{r61}.
    \item Ekermo defines win-win as “the theoretical possibility of finding mutually beneficial solutions for economy and environment” \cite{r62}.
    \item Smith  et al. define win-win as “the idea that one person's success is not achieved at the expense or exclusion of the success of others” \cite{r63}.
    \item Engler defines win-win as “a fair distribution of the efforts of the collaboration and the results” \cite{r64}
    \item Moon and Dathe-Douglass defines win-win as “the only rational way for a leader to think” \cite{r65}.
    \item Willing et al. define win-win as “the approach that seeks a mutually beneficial outcome, resulting in mutual cooperation and joint commitment in its implementation” \cite{r66}.
    \item Blount defines win-win as “the warm blanket of delusion where your commission check and your company's profits curl up to die” \cite{r67}.
    \item Dor defines win-win as “the art of winning while letting the other side think that they have won as well” \cite{r80}.
    \item And recently, Zhang et al.  define win-win as  “the realization of maximizing the interests of both sides, which is a harmonious development with mutual benefits” \cite{r60}.
    
\end{itemize}

The importance of the win-win notion is evidenced from its widespread applicability in almost all study fields, giving feasible answers to a vast array of challenges. Furthermore, a great number of scholars are praising the concept by giving it fancy titles based on the application, such as the key \cite{r71}, the solution \cite{r69}, the optimal choice \cite{r68}, the best model \cite{r70}, the cornerstone \cite{r78}, the ultimate goal \cite{r72}, the only way \cite{r73}, the final purpose \cite{r74}, the logical force \cite{r75}, the ideal way \cite{r76}, the ideal solution \cite{r77}, the objective \cite{r79}, etc. Moreover, by the time of writing this paper, searching the Google Scholar engine for the word “win-win” returned 616,000 results, while searching the Google engine returned 68,400,000, This is an indicator that the concept is widely used, whether in academia or in general.

Scholars’ differing opinions on win–win have resulted in a variety of explanations and definitions. Because of this difficulty, defining win–win as a rule of thumb is challenging. For example, Eshun et al. \cite{r49} defined win–win in Public–private partnership as a situation in which both the private and public sectors are equally and actively involved in the formulation of optimal and equitable risk assessment and allocation. While some scholars argue that negotiators with a fixed-pie bias consistently fail to attain optimal distributions because they do not seek out win–win solutions and are content with a just acceptable compromise \cite{r52,r53,r54}. On the contrary, Nalis et al. argue that such views might be incorrect, implying that there is a feasible solution that people just do not notice. As a result, it is crucial to figure out which personality qualities and circumstance conditions make developing such solutions easier \cite{r51}. Similarly, Lute stated that understanding the trends with the various management styles is extremely beneficial in building and sustaining a win-win environment, thus personal behaviour is attributed to the success of a win-win situation \cite{r50}. 

Because there are differing viewpoints, ambiguity, or at the very least misinterpretation of the notion of win-win, several researchers have worked to improve it. As an example, 
Fisher and Ury \cite{r47} presented a new form of win-win negotiation referred to as “principled negotiation”. They stated that effective negotiations promote collaboration toward a common objective. And they defined a number of stages to conducting principled negotiation, including:
\begin{itemize}
    \item Separate People From the Problem
    \item Focus on Interests, Not Positions
    \item Invent Options for Mutual Gain
    \item Use Objective Criteria
\end{itemize}

The authors of \cite{r1} proposed Goal-oriented balancing: happy–happy negotiations beyond win–win situations, suggesting that the term "win–win" is an oxymoron because it juxtaposes aspects that are incompatible (Only winners are possible because winning implies that someone will lose). As a result, they proposed the term "happy–happy situation" to better explain the heart of a successful negotiating process, suggesting that both parties should be satisfied. According to \cite{r1}, the concept "happy–happy" offers a number of advantages, including:

\begin{itemize}
    \item It isn't an oxymoron, thus it more accurately represents the desired situation.
    \item It isn't a competitive metaphor because it emphasizes fulfilment instead of competitiveness.
    \item It is relational rather than transactional in nature. 
    \item Because it depicts an emotion, it focuses on the process rather than an objective end. 
    \item It signifies that the negotiating process is ongoing and does not come to an end in a specific circumstance.
\end{itemize}

Another interesting study done by Paul Gruenbacher ~\cite{gruenbacher2000collaborative} employed win-win solution in the software development. More specifically, he used a methodology called easy win-win. In this methodology, the author suggests the involvement of stockholders in the negotiations to maximize the satisfactory level between the negotiators. In his methodology, the stockholders can be a part of processes like brainstorming, which can be done electronically and make them able to share their ideas simultaneously. This approach helps in simplify the involvement as well as enhance the interaction and communication of stockholders.

To the best of our knowledge, the only mathematical win-win model was proposed by Antonio Ruiz–Cortes and co-workers ~\cite{ruiz2002using}. They suggest an improvement on the traditional win-win in the requirement engineering stage of a given project. Their improvement was based on a mathematical representation of the requirements using the set theory and make constraints associated with each win condition in the requirement space.  The improvement is applicable in product–oriented contexts and where it can be many stockholders in the projects. The main benefit of their proposal is that it removes the conflicts before the negotiation starts in order to maximize the satisfaction for all the involved parties.

During our review of the literature on the win-win concept, we came across findings that contradict each other, such as the importance of teaching win-win \cite{r45} and the Myth of the Win-Win \cite{r46}. Furthermore, in terms of win-win, all prior research and definitions employed ambiguous or Fuzzy terms such as outcomes, most, success, efficiency, utility, fairness, beneficial, maximizing, etc. Knowing that there is no universal or absolute measure of happiness, fairness, success, etc., we claim that the concept of win-win is ambiguous or at least Fuzzy enough to produce such an inconsistency since it is not formally and accurately modelled. As a result, we are motivated to develop a new win-win model based on Fuzzy logic, which we refer to as Fuzzy win-win.

In a typical win-win situation, both parties believe they are winners; however, this is only partially accurate; in most circumstances, one side wins more, which is why we propose our Fuzzy win-win model, which attempts to quantify the winning situation for each party.

\section{The Proposed Fuzzy win-win model}

Almost all the models proposed in the literature were attempting to ensure satisfaction between different parties. However, no model answers the question that how much each party benefits from the deal or the negotiations? The Fuzzy win-win approach is based on the well-known Fuzzy logic, and it is proposed here to determine the winning proportion of each negotiator.
Fuzzy logic is first proposed by Lotfi Aliasker Zadeh in 1965 \cite{r84}, and since then it has been used in fields as diverse as Economics, Statistics, Finance, Management, Engineering, etc. \cite{r81, r82, r83, r85, r86, r87,r88,r89,r90,r91,r92, r93,r94,r95,r96,r97,ex1,ex2,ex3,ex4,ex5,ex6,ex7}.

Fuzzy logic uses membership functions to determine the percent of win for each party. To simplify the illustration of the proposed model, we consider two parties of the negotiation. The first one is the seller and the second one is the buyer. Let us assume that both are negotiating on selling/buying of a product, with the following parameters:

\begin{itemize}
    \item Product initial cost price (CP).
    \item The negotiation results in selling the product in price P.
    \item The seller wanted price (SP), normally this price is the fair market price of the product measured at the same time of the negotiated deal.  

\end{itemize}

Typically, $SP > CP$ therefore, and according to the win-win, if $P > CP$ then the seller is consider as a winner. At the same time, if $P < SP$ then the buyer is considered as a winner. However, if $P > CP \land P < SP$ then both parties are considered winners regardless P. And this is exactly the dilemma of the win-win, as both of the seller and buyer do not win the same; one of them might win more, and therefore benefits more from the deal. 
The previous aspects of the proposed Fuzzy win-win model can be quantitatively expressed using membership functions in order to quantify the quantity; or, to be more specific, the percentage of winning for each party as follows:

\begin{equation}
Seller(P) = \left\{
        \begin{array}{lll}
            0\% & \quad P < CP \\
            100\% & \quad P > SP \\
            \frac{P-CP}{SP-CP}\% & otherwise
        \end{array}
    \right.
\end{equation}

where Seller(P) is a function of P, which outputs the percentage of winning of the seller, and

\begin{equation}
Buyer(P) = \left\{
        \begin{array}{lll}
            100\% & \quad P < CP \\
            0\% & \quad P > SP \\
            \frac{SP-P}{SP-CP}\% & otherwise
        \end{array}
    \right.
\end{equation}

\noindent where Buyer(P) is a function of P, which outputs the percentage of winning of the buyer. 

Obviously, the Fuzzy win-win model has two membership functions. One is dedicated for the seller, and the other is dedicated for the buyer. Both functions can determine the winning percentage for each party.
Figure 1 shows a simple deal where the seller function is presented as blue dotted function, while the buyer function in dotted red.

\begin{figure}[ht]
    \centering
    \includegraphics{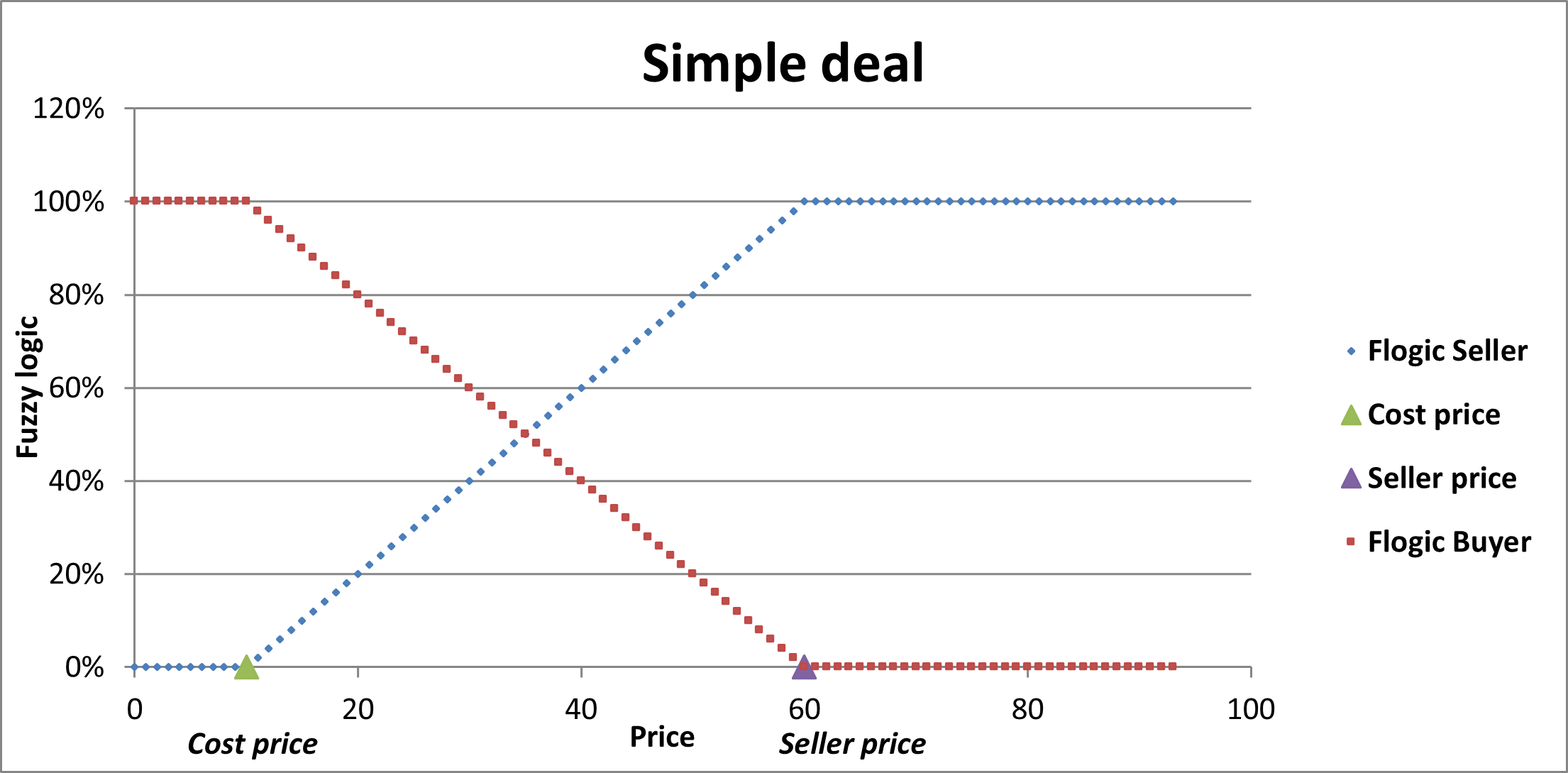}
    \caption{Illustration of the membership function using a simple deal example. In this deal the cost price of the product is 10 and the seller price is 60 regardless the currency.}
    \label{fig:fig1}
\end{figure}

According to the previous Equations (1 and 2) and the curves in Figure 1, we can calculate the percentage of winning for each party. For example, the buyer is considered 100\% winning if they could secure the deal with P $\leq$ CP. However, this rate decreases as P increases, until it reaches 0\% if the buyer paid SP or more,  because if the buyer wants to sell what he just bought they will most likely get  only what they paid without any profit.  On the other hand, the seller is considered 0\% if they secured the deal with P $\leq$ CP, because they got only the cost price or less with no profit gained from the deal. However, this rate increases as P increases, until it reaches 100\% if the seller could get his wanted price SP or more.

It is worth mentioning that the previous curves in Figure 1 show 3 distinct areas that describe 3 different situations, namely: 

\begin{enumerate}
    \item Lose-win situation, starting from P=0 until P=CP.
    \item Fuzzy win-win situation, starting from P=CP until P=SP, the so-called zone of possible agreement (ZOPA).
    \item Win-lose situation, starting from P=SP until P= infinity.
\end{enumerate}

However, in this work we focus on the Fuzzy win-win situation, because the goal of this paper is to quantify the win-win situation alone.

The inverse functions of Equations (1 and 2) are also important for our model, because by using them, we can obtain P given a percentage of winning of any party, and can be formulated as follows 

\begin{equation}
P_s = \left\{
        \begin{array}{lll}
            CP & \quad SWP=0\% \\
            SP & \quad SWP=100\% \\
            SWP * (SP-CP)+CP & otherwise
        \end{array}
    \right.
\end{equation}

where $P_s$ is the price that the seller should secure the deal with, if they want to achieve a specific seller winning percentage (SWP) out of the deal, and

\begin{equation}
P_b = \left\{
        \begin{array}{lll}
            CP & \quad BWP = 100\% \\
            SP & \quad BWP = 0\% \\
           SP-BWP*(SP-CP) & otherwise
        \end{array}
    \right.
\end{equation}

where $P_b$ is the price that the Buyer should secure the deal with, if they want to achieve a specific buyer winning percentage (BWP) out of the deal. Figure 2 depicts the curves of the inverse functions of the Fuzzy win-win model.

\begin{figure}[ht]
    \centering
    \includegraphics{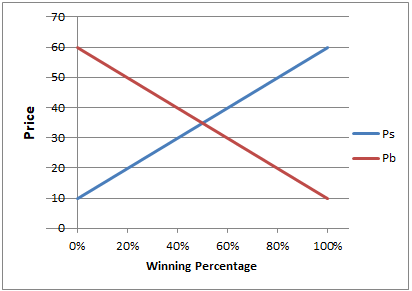}
    \caption{Illustration of the Fuzzy win-win inverse functions (Equations 3 and 4) based on the same example in Figure 1, where CP=10 and SP=60 regardless the currency.}
    \label{fig:fig2}
\end{figure}

As can be seen from Figure 2, the Domain of both functions Ps and Pb is defined for all real values in the range of $[0,1]$, hence both functions can take any winning percentage from 0\% to 100\%. And the Range of both functions is defined for all real values in the range of $[CP, SP]$; in our example 10 and 60. Therefore, both functions can suggest a specific price given a winning percentage.

\section{Results and discussion}

In order to apply the proposed Fuzzy win-win system to solve a specific problem, we used a toy example for an imaginary simple deal, and 3 real-world cases. Namely, Iraqi-Jordanian current oil deal, and the international iron ore negotiations (2005-2009).

\subsection{Toy Example}
Assume that the CEOs of two corporations, company C1, and company C2, are negotiating the purchase of some of the latter's shares. The current price of the share is \$ 33. However, it is expected- for any reason to be doubled soon. How we can use the Fuzzy win-win model to help any CEO in this negotiation?
According to the normal win-win concept, both of the CEOs will be happy and winners if they secured the deal with any price in the range of [\$33, \$66]. However, The Proposed Fuzzy win-win model is more precise, and can answer two types of questions for each party:
\\
Q1) if the share is sold for P; what is the winning percentage for each of C1 and C2?
\\
Q2) if the CEO of C2 wants to achieve a specific winning percentage out of this deal; in what price should C2 sell its share? 
\\
And similarly, if the CEO of C1 wants to achieve a specific winning percentage out of this deal; in what price should C1 Buy the share of C2?
\\
By applying the proposed Fuzzy win-win model (Equations 1-4) on the data of our toy example we get Table 1.

\begin{table}
\centering
\small
\caption{A toy example illustrates the calculation wining percentages produced using the proposed method.}
\begin{tabular}{|l|c|c|c|c|} 
\hline
\textbf{ Share price } & \textbf{ C2-SWP } & \textbf{ C1-BWP } & \textbf{ Ps } & \textbf{ Pb } \\ 
\hline
33 & 0\% & 100\% & 33 & 33 \\ 
\hline

35 & 6\% & 94\% & 35 & 35 \\ 
\hline

37 & 12\% & 88\% & 37 & 37 \\ 
\hline

39 & 18\% & 82\% & 39 & 39 \\ 
\hline

41 & 24\% & 76\% & 41 & 41 \\ 
\hline

43 & 30\% & 70\% & 43 & 43 \\ 
\hline

45 & 36\% & 64\% & 45 & 45 \\ 
\hline

47 & 42\% & 58\% & 47 & 47 \\ 
\hline

49 & 48\% & 52\% & 49 & 49 \\ 
\hline

51 & 55\% & 45\% & 51 & 51 \\ 
\hline

53 & 61\% & 39\% & 53 & 53 \\ 
\hline

55 & 67\% & 33\% & 55 & 55 \\ 
\hline

57 & 73\% & 27\% & 57 & 57 \\ 
\hline

59 & 79\% & 21\% & 59 & 59 \\ 
\hline

61 & 85\% & 15\% & 61 & 61 \\ 
\hline

63 & 91\% & 9\% & 63 & 63 \\ 
\hline

65 & 97\% & 3\% & 65 & 65 \\ 
\hline

\end{tabular}

\end{table}

Table 1 shows discrete data, that is because the share price in the table has been increased by one for illustrative purposes. The Fuzzy win-win, on the other hand, is continuous by nature, as inherited from Fuzzy logic. If we employ the model's equations, we can get real values. We can answer any of the previous questions for C1 or C2 in either case. For example, if the deal was secured on \$40, the SWP for C2 is 21\%, and the BWP for C1 is 79\%. Both percentages are complements of each other, this is because the equations used are linear, these functions and curves are subject to altering and tuning by the users of the system. On the other hand, if CEO of C2 came to the negotiations with 40\% SWP in mind, they need to sell their share in \$46.2.
If taken literally, the win-win concept could be equivalent to the Fuzzy win-win when both parties achieve 50\% winning percentage. For this toy example, the share needs to be sold for \$49.5. It is worth noting that Fuzzy win-win is not designed for predicting winning/losing situations, however, it is based on current information and given data, if we assumed that the price will be doubled, then this is a fact for the Fuzzy win-win, regardless of its truthiness/falseness since it works based only on the given data.

\subsection{Real-world case 1: Iraqi-Jordanian current oil deal}

In case number one, we'll look at the oil deal struck between the governments of Iraq and Jordan between September 2019 and November 2020 \cite{oil}. The Jordanian government has agreed to buy a barrel of oil crude for \$16 less than the Brent price. The barrel's production cost was \$12.57. Table 2 shows the price information and the Fuzzy win-win calculations.

\begin{table}[ht!]
\centering
\caption{Information about the oil deal between Iraq and Jordan, with the winning percentages which calculated using the proposed method}
\begin{adjustbox}{width=\textwidth}
\begin{tabular}{|l|c|c|c|c|c|c|c|} 
\hline
\textbf{ Date } & \textbf{ Production cost } & \textbf{ Brent international price } & \textbf{ Jordan Buying Price } & \textbf{ Iraq SWP } & \textbf{ Jordan BWP } & \textbf{ Iraq Wins } & \textbf{ Jordan Wins } \\ 
\hline
\textbf{ Dec 30 } & 10.57 & 68.44 & 52.44 & 72\% & 28\% & X & ~ \\ 
\hline
\textbf{ Jan 6 } & 10.57 & 68.91 & 52.91 & 73\% & 27\% & X & ~ \\ 
\hline
\textbf{ Jan 13 } & 10.57 & 64.2 & 48.2 & 70\% & 30\% & X & ~ \\ 
\hline
\textbf{ Jan 21 } & 10.57 & 64.59 & 48.59 & 70\% & 30\% & X & ~ \\ 
\hline
\textbf{ Jan 27 } & 10.57 & 59.32 & 43.32 & 67\% & 33\% & X & ~ \\ 
\hline
\textbf{ Feb 3 } & 10.57 & 54.45 & 38.45 & 64\% & 36\% & X & ~ \\ 
\hline
\textbf{ Feb 10 } & 10.57 & 53.27 & 37.27 & 63\% & 37\% & X & ~ \\ 
\hline
\textbf{ Feb 18 } & 10.57 & 57.75 & 41.75 & 66\% & 34\% & X & ~ \\ 
\hline
\textbf{ Feb 24 } & 10.57 & 56.3 & 40.3 & 65\% & 35\% & X & ~ \\ 
\hline
\textbf{ Mar 2 } & 10.57 & 51.9 & 35.9 & 61\% & 39\% & X & ~ \\ 
\hline
\textbf{ Mar 6 } & 10.57 & 45.27 & 29.27 & 54\% & 46\% & X & ~ \\ 
\hline
\textbf{ Mar 10 } & 10.57 & 37.22 & 21.22 & 40\% & 60\% & ~ & X \\ 
\hline
\textbf{ Mar 16 } & 10.57 & 30.05 & 14.05 & 18\% & 82\% & ~ & X \\ 
\hline
\textbf{ Mar 24 } & 10.57 & 27.15 & 11.15 & 3\% & 97\% & ~ & X \\ 
\hline
\textbf{ Mar 30 } & 10.57 & 22.76 & 6.76 & 0\% & 100\% & ~ & X \\ 
\hline
\textbf{ Apr 7 } & 10.57 & 31.87 & 15.87 & 25\% & 75\% & ~ & X \\ 
\hline
\textbf{ Apr 14 } & 10.57 & 29.6 & 13.6 & 16\% & 84\% & ~ & X \\ 
\hline
\textbf{ Apr 20 } & 10.57 & 25.57 & 9.57 & 0\% & 100\% & ~ & X \\ 
\hline
\textbf{ Apr 28 } & 10.57 & 20.46 & 4.46 & 0\% & 100\% & ~ & X \\ 
\hline
\textbf{ May 4 } & 10.57 & 27.2 & 11.2 & 4\% & 96\% & ~ & X \\ 
\hline
\textbf{ May 11 } & 10.57 & 29.63 & 13.63 & 16\% & 84\% & ~ & X \\ 
\hline
\textbf{ May 18 } & 10.57 & 34.81 & 18.81 & 34\% & 66\% & ~ & X \\ 
\hline
\textbf{ May 26 } & 10.57 & 35.53 & 19.53 & 36\% & 64\% & ~ & X \\ 
\hline

\end{tabular}

\end{adjustbox}
\end{table}


\begin{table}
\centering
\caption{Information about the oil deal between Iraq and Jordan, with the winning percentages which calculated using the proposed method}
\begin{adjustbox}{width=\textwidth}
\begin{tabular}{|l|c|c|c|c|c|c|c|} 
\hline
\textbf{ Date } & \textbf{ Production cost } & \textbf{ Brent international price } & \textbf{ Jordan Buying Price } & \textbf{ Iraq SWP } & \textbf{ Jordan BWP } & \textbf{ Iraq Wins } & \textbf{ Jordan Wins } \\ 
\hline

\textbf{ Jun 1 } & 10.57 & 38.32 & 22.32 & 42\% & 58\% & ~ & X \\ 
\hline
\textbf{ Jun 8 } & 10.57 & 40.8 & 24.8 & 47\% & 53\% & ~ & X \\ 
\hline
\textbf{ Jun 15 } & 10.57 & 39.72 & 23.72 & 45\% & 55\% & ~ & X \\ 
\hline
\textbf{ Jun 22 } & 10.57 & 43.08 & 27.08 & 51\% & 49\% & X & ~ \\ 
\hline
\textbf{ Jun 29 } & 10.57 & 41.71 & 25.71 & 49\% & 51\% & ~ & X \\ 
\hline
\textbf{ Jul 6 } & 10.57 & 43.1 & 27.1 & 51\% & 49\% & X & ~ \\ 
\hline
\textbf{ Jul 13 } & 10.57 & 42.72 & 26.72 & 50\% & 50\% & X & X \\ 
\hline
\textbf{ Jul 20 } & 10.57 & 43.28 & 27.28 & 51\% & 49\% & X & ~ \\ 
\hline
\textbf{ Jul 27 } & 10.57 & 43.41 & 27.41 & 51\% & 49\% & X & ~ \\ 
\hline
\textbf{ Aug 3 } & 10.57 & 44.15 & 28.15 & 52\% & 48\% & X & ~ \\ 
\hline
\textbf{ Aug 10 } & 10.57 & 44.99 & 28.99 & 54\% & 46\% & X & ~ \\ 
\hline
\textbf{ Aug 17 } & 10.57 & 45.37 & 29.37 & 54\% & 46\% & X & ~ \\ 
\hline
\textbf{ Aug 24 } & 10.57 & 45.13 & 29.13 & 54\% & 46\% & X & ~ \\ 
\hline
\textbf{ Aug 31 } & 10.57 & 42.61 & 26.61 & 50\% & 50\% & X & X \\ 
\hline
\textbf{ Sep 8 } & 10.57 & 39.78 & 23.78 & 45\% & 55\% & ~ & X \\ 
\hline
\textbf{ Sep 14 } & 10.57 & 39.61 & 23.61 & 45\% & 55\% & ~ & X \\ 
\hline
\textbf{ Sep 21 } & 10.57 & 41.44 & 25.44 & 48\% & 52\% & ~ & X \\ 
\hline
\textbf{ Sep 28 } & 10.57 & 42.43 & 26.43 & 50\% & 50\% & X & X \\ 
\hline
\textbf{ Oct 5 } & 10.57 & 41.29 & 25.29 & 48\% & 52\% & ~ & X \\ 
\hline
\textbf{ Oct 12 } & 10.57 & 41.72 & 25.72 & 49\% & 51\% & ~ & X \\ 
\hline
\textbf{ Oct 19 } & 10.57 & 42.62 & 26.62 & 50\% & 50\% & X & X \\ 
\hline
\textbf{ Oct 26 } & 10.57 & 40.46 & 24.46 & 46\% & 54\% & ~ & X \\ 
\hline
\textbf{ Nov 2 } & 10.57 & 38.97 & 22.97 & 44\% & 56\% & ~ & X \\ 
\hline
\textbf{ Nov 9 } & 10.57 & 42.4 & 26.4 & 50\% & 50\% & X & X \\ 
\hline
\textbf{ Nov 16 } & 10.57 & 43.62 & 27.62 & 52\% & 48\% & X & ~ \\ 
\hline
\textbf{ Nov 23 } & 10.57 & 46.06 & 30.06 & 55\% & 45\% & X & ~ \\ 
\hline
\textbf{ Nov 30 } & 10.57 & 47.59 & 31.59 & 57\% & 43\% & X & ~ \\ 
\hline
\textbf{ AVG. } & ~ & ~ & ~ & 45\% & 55\% & 27 & 28 \\
\hline
\end{tabular}

\end{adjustbox}
\end{table}

By applying our equations, we get the seller win percent SWP, Iraq, and the buyer win percent BWP, Jordan. These percentages show how many times, and at which prices each country achieved more winning percentages. As can be clearly seen in Table 2, the deal is almost fair, however, there is more winning percentages for Jordan side. Jordan won more in 28 cases, while Iraq won 27 times. The average of winning percentages is 45\% and 55\% for Iraq and Jordan, respectively. The win percentages were perfectly equal in 5 cases, according to the data in the table. In 3 cases out of 50, Jordan achieved 100\% win percent. Figure 3 shows the winning percentages curves for Iraq and Jordan.

\begin{figure}[ht]
    \centering
    \includegraphics{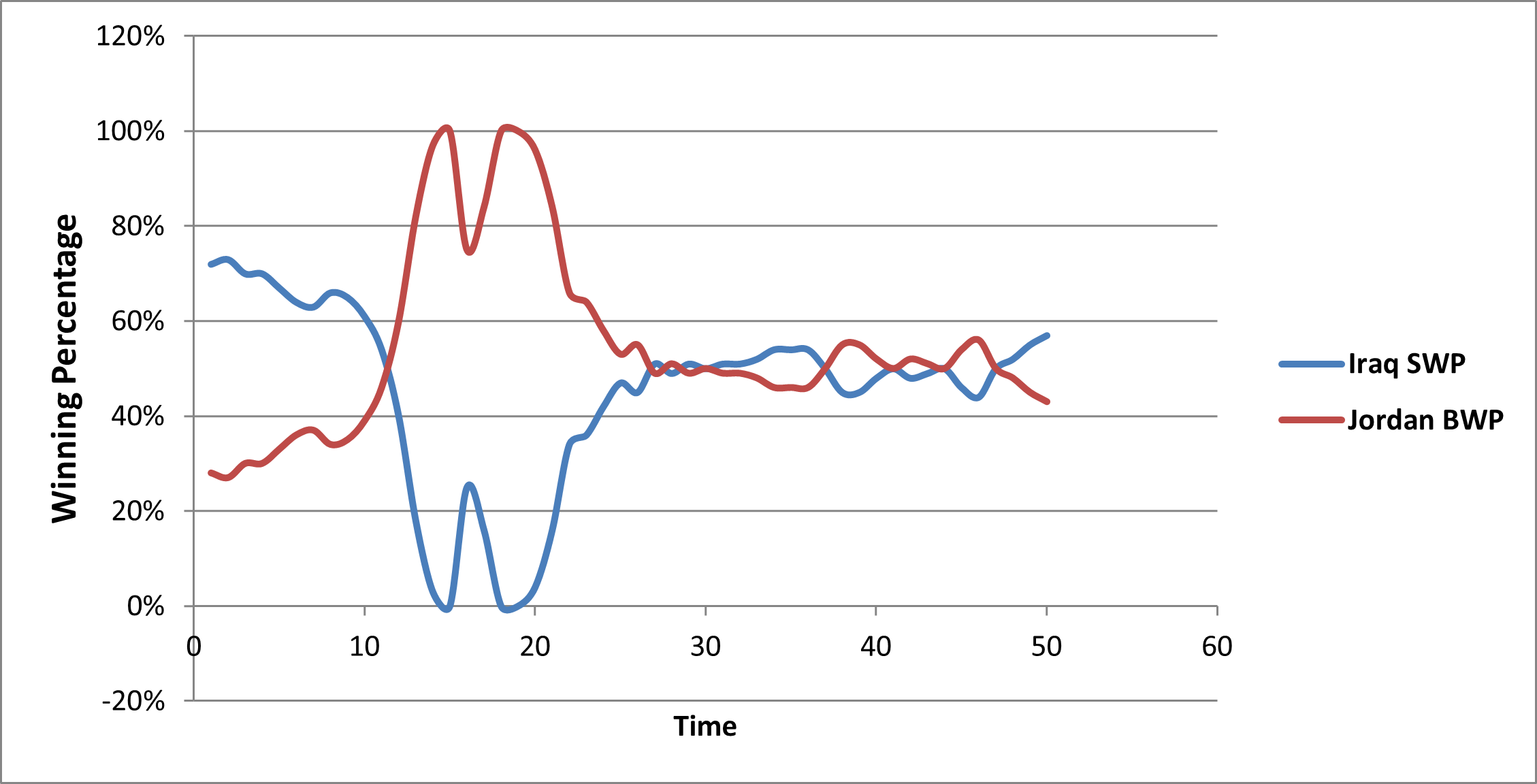}
    \caption{Winning percentages curves for Jordan and Iraq in the oil deal. The graph displays the winning rates as a function of time (data records).}
    \label{fig:fig3}
\end{figure}

\subsection{Real- world case 2: International Iron Ore Negotiations (2005-2009) }

This example discusses the negotiations on iron ore that took place between the major mining companies (sellers), including Companhia Vale do Rio Doce (known as both CVRD and Vale) located in Brazil, Rio Tinto Group located in Australia, and the Anglo-Australian firm BHP Billiton (BHP) also located in Australia. The example covers the negotiations in the years between 2005 and 2009. The buyers were the big steel companies located in countries from Asia, Europe, and North America \cite{r98}.

In 2005, the first offer announced by BHP was 50\% price increase, Vale, one of the largest Iron Ore producers, announced that it had received an offer with a 90\% price rise on the older market price. After negotiations, Nippon (one of the largest Iron Ore buyers) and Vale eventually agreed on a 71.5\% increase a month later.

In 2006, mining firms proposed 10\% to 20\% price increase, Vale proposed a 24\% increase. After arduous negotiations, Vale and ThyssenKrupp announced that they agreed on a 19\% price increase.

In 2007, mining firms proposed 5\% to 10\% price increase, Baosteel and Vale announced that they agreed on a 9.5\% price increase.

In 2008, Vale asked for a 70 percent price increase, Nippon and Posco announced an agreement with Vale, calling for a two-tiered price increase of 65\% in the price of southern ore, and a 71\% increase in the price of northern ore. Here we have two deals, one was secured with higher increase than what the seller asked for, and one with less increase that what the seller asked for. 

In 2009, as a result of the global financial crisis, the spot price of iron ore began to decline, and it soon fell below the agreed-upon negotiation price. Therefore, some of the buyers refused to import Iron ore breaching their contracts. As a response, the major buyers of the ore proposed reduced prices; for example, Baosteel proposed 45\% decrease of the price. Nippon and Rio Tinto reached agreement at price decreases of 33\%, and 44\% price decrease of the other type of the ore. 

Before applying the proposed Fuzzy win-win model on these deals, we need to define its parameters as follows:

\begin{itemize}
    \item 	2005 negotiations: CP = 50\% price increase, SP=90\% price increase, and P= 71.5\% price increase.
\item 		2006 negotiations: CP = 10\% price increase, SP=24\% price increase, and P= 19\% price increase.
\item 		2007 negotiations: CP = 5\% price increase, SP=10\% price increase, and P= 9.5\% price increase.
\item 		2008 negotiations: CP = 10\% price increase (as we assumed that the buyers’ negotiator came with 10\% increase in mind, like that of the previous year), SP=70\% price increase, and P= 65\% price increase. And for the other type of the ore; P=71\% price increase.
\item 		2009 negotiations: CP = 0\% price decrease (as we assumed that the seller’s negotiator came with 0\% decrease in mind, as they used to increase the price of the ore in the previous years), SP=45\% price decrease, and P= 33\% price decrease, and P= 44\% price decrease of the other type of the ore.

\end{itemize}

After applying the proposed Fuzzy win-win model we get the data in Table 4.

\begin{table}
\centering
\caption{Fuzzy win-win results of the international Iron Ore negotiations (2005-2009)}
\small
\begin{tabular}{|l|l|l|l|l|l|l|l|} 
\hline
\textbf{ Year } & \textbf{ CP } & \textbf{ SP } & \textbf{ P } & \textbf{ SWP } & \textbf{ BWP } & \textbf{ Sellers  win more } & \textbf{ Buyers  win more } \\ 
\hline
\textbf{ 2005 } & 50\% & 90\% & 71.5\% & 54\% & 46\% & X &  \\ 
\hline
\textbf{ 2006 } & 10\% & 24\% & 19.0\% & 64\% & 36\% & X &  \\ 
\hline
\textbf{ 2007 } & 5\% & 10\% & 9.5\% & 90\% & 10\% & X &  \\ 
\hline
\textbf{ 2008 Southern ore } & 10\% & 70\% & 65.0\% & 92\% & 8\% & X &  \\ 
\hline
\textbf{ 2008 Northern~ ore } & 10\% & 70\% & 71.0\% & 100\% & 0\% & X &  \\ 
\hline
\textbf{ 2009 Southern ore } & 0\% & 45\% & 33.0\% & 27\% & 73\% &  & X \\ 
\hline
\textbf{ 2009 Northern ore } & 0\% & 45\% & 44.0\% & 2\% & 98\% &  & X \\ 
\hline
\textbf{ Average } & 12\% & 51\% & 45\% & 61\% & 39\% & 5 & 2 \\
\hline
\end{tabular}
\end{table}

\begin{figure}
    \centering
    \includegraphics{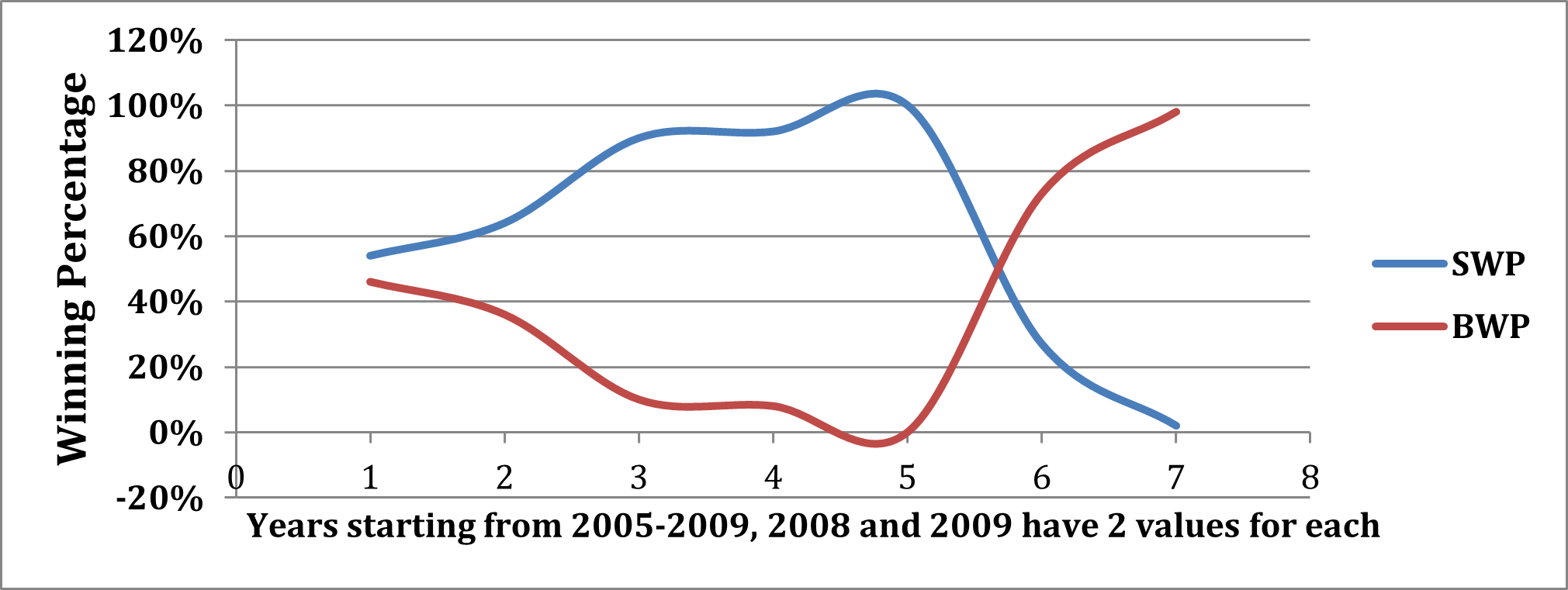}
    \caption{Sellers and buyers winning percentages in the ore negotiations}
    \label{fig:fig5}
\end{figure}

As can be seen from Table 4, the sellers won the negotiations 5 times on a row, from 2005-2008. In 2009 the buyers’ winning percent were 27\% and 2\% for Southern ore and Northern ore, respectively. The average winning percent for the sellers was 61\%, while it was 39\% for the buyers, for the 5 years. The sellers befitted more from the first 5 deals because there was a great demand on the ore during the period (2005-2008). However after the global financial crisis of 2007–2008 hit, the buyers become winning more according to the proposed Fuzzy logic win-win.

The proposed fuzzy win-win can be utilized in game theory to quantify a large number of scenarios, or more precisely, in any cases where the classic win-win term is applied \cite{r6,r7,r8,r9,r10,r35}. For example, in a Chess game, Fuzzy win-win can be used to quantify the amount of winning when a novice player loses against a grandmaster after a large number of moves; yes, losing Chess for a grandmaster can be considered a wining situation for a novice payer, particularly when playing a good and long game, as it can be considered good practice and learning experience.

Before we can apply the proposed Fuzzy win-win approach to the previous Novice-Grandmaster scenario, we must first define the minimum and maximum number of moves that both players must make. Perhaps the average number of moves in Chess is the best number we can choose, which is equal to 38 moves according to \cite{hsu1989large}, and let us assume that the maximum number of moves is double that number. These numbers can be changed depending on the end user's preferences and definition of a win-win situation, as well as the players' ranking.

\begin{figure}
    \centering
    \includegraphics{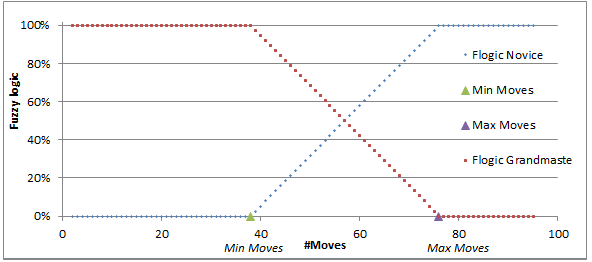}
    \caption{Fuzzy win-win model for Novice-Grandmaster Chess scenario.}
    \label{fig:chess}
\end{figure}

Having said that, according to the Fuzzy win-win model, a novice player will start winning after 38 moves, and if he/she loses the game with less than 38 moves, he/she will be considered a 100 percent winner if he/she plays more than 76 moves, whereas a grandmaster will be considered a 100 percent winner if he/she wins the game playing less than 38 moves, and a loser if he/she wins playing more than 76 moves. The Fuzzy area is located between 38 and 76 moves for both players. Figure \ref{fig:chess} depicts the Fuzzy win-win model for such a scenario.

As can be seen from Figure \ref{fig:chess}, the Fuzzy win-win can answer questions such as:
Q1) If the Game ended with 50 moves, what is the Fuzzy win-win situation for each of the Novice and Grandmaster players? The model response: Novice’s win-win is 32\%, whereas Grandmaster’s win-win is 68\%.
Q2) If the Novice player wants to achieve 60\% wining out of his/her game, in what number of moves should he/she lose the game? The model response: Novice’s Number of moves is 61.
Q3) If the Grandmaster wants to achieve 60\% wining out of his/her game, in what number of moves should he/she win the game? The model response: Grandmaster’s Number of moves is 53.
Solving Chess puzzles is another area where the proposed Fuzzy win-win approach can be put to good use. For example, a 100 percent winning situation could be defined as solving a puzzle by making all of the optimal/correct moves, and the winning percentage would drop to 0 percent if the player made no optimal/correct moves. This could be useful when assessing a player rank of Chess puzzles; if a Chess puzzle is partially solved, the player rank should be reduced by a fraction of points rather than by an arbitrarily large number of points.

We study the implementation of the proposed Fuzzy win-win in several research areas, ranging from business, economy, and politics to the fun stuff of game theory, as a result of the prior discussion. However, the proposed approach's application is not restricted to these sectors; it can be utilized in any circumstance where the traditional win-win concept applies.

\section{Conclusion}
In this work we proposed a new mathematical Fuzzy logic-based model that helps in quantifying the win-win situation. The model is based on an artificial intelligence knowledge representation approach called Fuzzy logic. The main advantage of the proposed Fuzzy win-win model is that it can determine the win percentages for each party, and inversely, it can provide the value to negotiate for given a winning percentage.

The model is tested on experimental real-life negotiation cases, in which there were sellers and buyers negotiating on a product or service, each of them wants to achieve the highest winning percentage. The findings suggest that the proposed Fuzzy win-win model can assist negotiators in getting the best feasible deal by providing quantified winning percentages for each party, something the traditional win-win  cannot provide. Furthermore, the proposed Fuzzy win-win approach has been shown to be effective in game theory for win-win situations, and two scenarios in the Chess game have been explored to demonstrate the potential of the proposed Fuzzy win-win approach.

In its current form, the Fuzzy win-win model can be applied to any negotiation-like circumstance where both sides share the same benefit type (x-axis), such as money, enrichment rates, water amount, energy, and so on. It cannot, however, provide answers for parties with opposing viewpoints on benefit, such as factory owners and the environment, where the first party considers money and the second party or representative considers the amount of gas emitted. To handle such problems, a modified form of the Fuzzy win-win can be utilized, which consists of independent curves/functions for each party. In our future work, we will address such enhancements, among others like the win-win-win situation.

\section*{References}

\bibliography{FWW}

\begin{thebibliography}{100}
\bibitem{r24}
A.~Lajeunesse, Finding'win-win'--china's arctic policy and what it means for
  canada, The School of Public Policy Publications.
\end{thebibliography}

\end{document}